\documentclass{article}
\usepackage{iclr2027_conference,times}
\usepackage[T1]{fontenc}
\iclrfinalcopy %

\usepackage{amsmath,amsfonts,bm}

\def\eqref#1{equation~\ref{#1}}

\def\1{\bm{1}}

\DeclareMathAlphabet{\mathsfit}{\encodingdefault}{\sfdefault}{m}{sl}
\SetMathAlphabet{\mathsfit}{bold}{\encodingdefault}{\sfdefault}{bx}{n}

\usepackage{hyperref}
\hypersetup{
  pdftitle={Visualizing Distribution Coverage in Generative Diffusion Models},
  pdfauthor={Yifei Wang, Xiaoyu Wu, Tsu-Jui Fu, Chen Chen, Liang-Chieh Chen, Zhe Gan, Chen Wei},
  pdfsubject={arXiv preprint},
  hidelinks
}
\usepackage{url}
\usepackage{amssymb}
\usepackage{graphicx}
\usepackage{booktabs}
\usepackage{xcolor}
\graphicspath{{figures/}}
\usepackage[section]{placeins} %
\usepackage{afterpage} %
\usepackage[font=small]{caption}
\makeatletter
\def\section{\@startsection {section}{1}{\z@}{-1.6ex plus
    -0.4ex minus -.2ex}{1.0ex plus 0.2ex
minus0.2ex}{\large\sc\raggedright}}
\def\subsection{\@startsection{subsection}{2}{\z@}{-1.4ex plus
-0.4ex minus -.2ex}{0.6ex plus .2ex}{\normalsize\sc\raggedright}}
\def\paragraph{\@startsection{paragraph}{4}{\z@}{1.0ex plus
0.4ex minus .2ex}{-1em}{\normalsize\bf}}
\g@addto@macro\normalsize{%
  \setlength{\abovedisplayskip}{5pt plus 2pt minus 2pt}%
  \setlength{\belowdisplayskip}{5pt plus 2pt minus 2pt}%
  \setlength{\abovedisplayshortskip}{2pt plus 1pt}%
  \setlength{\belowdisplayshortskip}{3pt plus 1pt minus 1pt}}
\makeatother
\usepackage{enumitem}
\setlist[itemize]{nosep,topsep=2pt,leftmargin=*}
\title{Visualizing Distribution Coverage in \\ Generative Diffusion Models}

\author{%
  Yifei Wang\textsuperscript{1}\;
  Xiaoyu Wu\textsuperscript{1}\;
  Tsu-Jui Fu\textsuperscript{2}\;
  Chen Chen\textsuperscript{2}
  \\[0.3em]
  \textbf{Liang-Chieh Chen\textsuperscript{2}\;
  Zhe Gan\textsuperscript{2}\;
  Chen Wei\textsuperscript{1}}
  \\[0.6em]
  \textsuperscript{1}Rice University
  \qquad
  \textsuperscript{2}Apple
}

\begin{document}

\maketitle
\fancyhead{}
\renewcommand{\headrulewidth}{0pt}

\begin{abstract}
Diffusion distillation is widely adopted to accelerate sampling, and the resulting few-step models are broadly believed to match or even surpass their multi-step teachers in generation. However, standard evaluations such as GenEval2 typically draw only one sample per prompt, so improved scores may fail to reveal losses in distribution coverage. We therefore revisit whether distilled models truly match their teachers beyond single-draw performance using \textbf{pass@$\mathbf{k}$}, which measures the probability that at least one of $k$ independent samples satisfies a quality criterion. At $k{=}1$, pass@$k$ reduces to standard single-draw evaluation. As $k$ grows, the curve reveals whether additional draws find genuinely different successes or merely revisit the same modes, directly exposing how broadly a model covers the space of valid outputs. We first show that classifier-free guidance (CFG), whose quality--coverage tradeoff is well established, is the clearest case: higher guidance improves pass@$1$, but its advantage shrinks and reverses at larger $k$. Applying pass@$k$ to few-step distilled models, we find the same tradeoff splits along training objectives: distribution-matching objectives concentrate the student's output distribution, boosting early-hit rates while eroding large-budget coverage, whereas consistency and trajectory-based objectives better preserve the teacher's coverage even at large $k$. We further show that this tradeoff extends to few-step causal video generation. Our findings reveal a previously overlooked cost of diffusion distillation: across both image and video generation, the choice of training objective fundamentally determines whether a few-step model inherits its teacher's distribution coverage or trades it away for single-draw quality.

\end{abstract}

\afterpage{%
\begin{figure}[t!]
\centering
\includegraphics[width=0.99\linewidth]{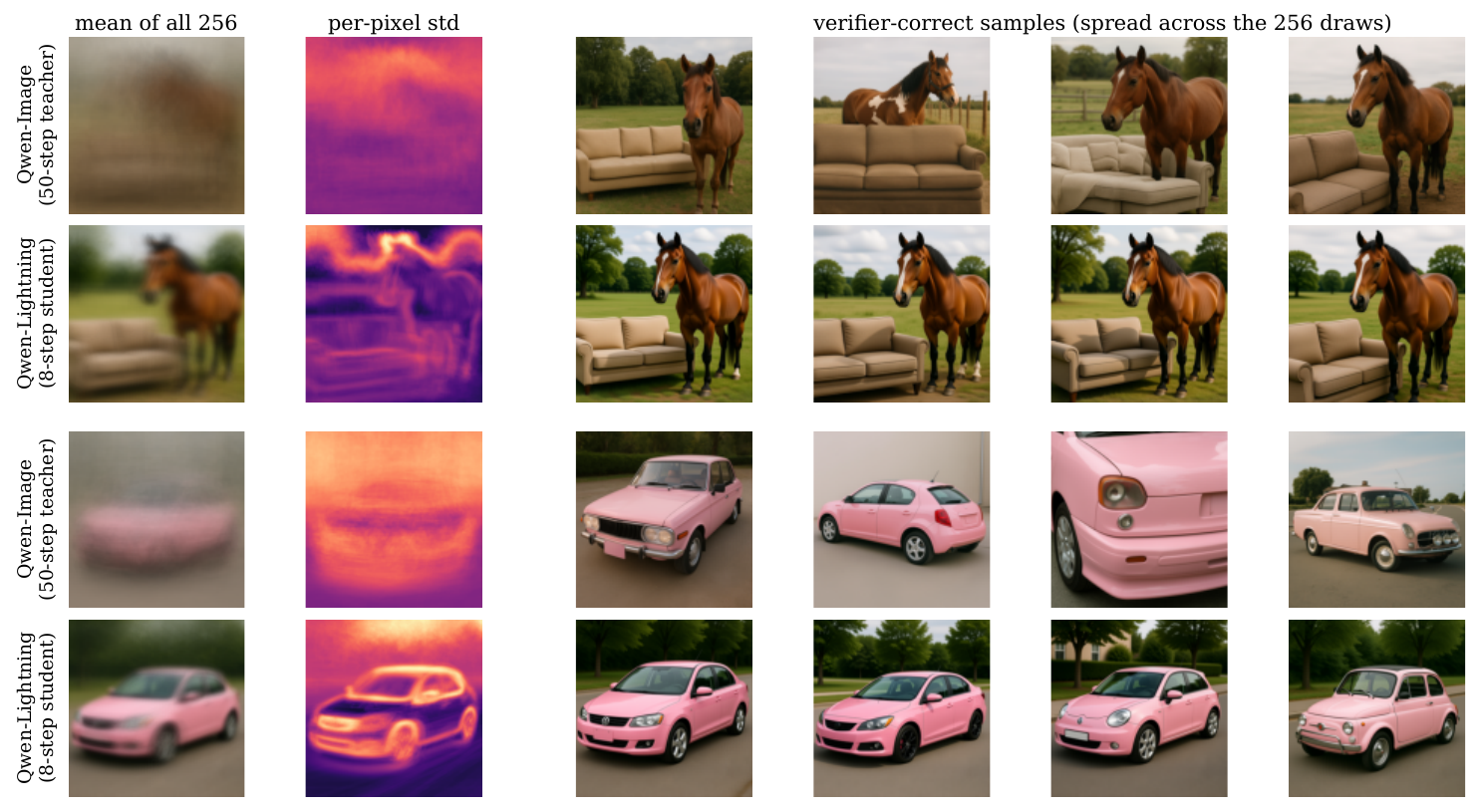}
\caption{256 independent samples from
Qwen-Image (50-step teacher) and its few-step student Qwen-Lightning for ``a
photo of a couch and a horse'' (top) and ``a photo of a pink car'' (bottom). Each row shows the
pixelwise mean of the 256 samples, the per-pixel standard deviation, and four samples. The student produces sharp individual samples but its lower standard deviation reveals a narrower output distribution: additional draws yield redundant variants rather than new valid compositions.}
\label{fig:teaser}
\end{figure}%
}
\section{Introduction}

Diffusion models~\citep{ho2020denoising,song2021denoising,song2021score} generate high-quality images and videos but require iterative denoising, making sampling slow and expensive. Distillation addresses this by training a student model to reproduce the teacher's output distribution in fewer steps, reducing inference cost by an order of magnitude or more while retaining the teacher's pretrained knowledge~\citep{luo2023latent,song2023consistency}. The resulting speedups, often 10--50$\times$ fewer function evaluations, have made distilled models the default choice for latency-constrained diffusion deployment~\citep{chen2025sana,sauer2024adversarial,zhou2025score,yin2024improved}.

The resulting few-step models are broadly believed to match or even surpass their multi-step teachers in generation quality. Indeed, on compositional benchmarks such as GenEval~\citep{ghosh2023geneval} and GenEval2~\citep{kamath2025geneval}, distilled models routinely match or exceed their teachers' scores~\citep{sauer2024adversarial,chen2025sana}. However, these benchmarks are typically evaluated with only one sample per prompt, so a model can score better simply by concentrating probability on a narrower set of high-scoring outputs, even as its coverage of other valid modes declines. In practice, distilled models are often observed to produce less varied outputs than their teachers, yet no established metric quantifies this coverage gap or identifies its source.

Fig.~\ref{fig:teaser} illustrates what these metrics miss. For two prompts, we draw 256 independent samples from a 50-step teacher and its few-step student. The student's pixel-wise standard deviation is visibly lower: it concentrates probability on fewer modes while the teacher spreads mass more broadly. Both models can produce individually convincing samples, yet the student's distribution is narrower. In practice, drawing multiple samples per prompt is the norm: a user may browse a batch to select a preferred composition, or a best-of-$N$ pipeline may score candidates and return the strongest. The value of this extra budget depends on each draw exploring a different valid interpretation of the prompt. A narrow distribution undercuts this: additional samples converge toward similar outputs, and the marginal gain from each extra draw shrinks.

To quantify this effect, we adopt pass@$k$ from code-generation evaluation~\citep{chen2021evaluating}. For each prompt, let $q_\theta(x)$ denote the probability that one sample satisfies a benchmark success criterion. Pass@$1$ reports the mean of $q_\theta(x)$; the pass@$k$ curve reveals how that success probability is distributed across prompts as the sampling budget grows. When a model's advantage at small $k$ shrinks or reverses at large $k$, it signals that the model has concentrated its distribution rather than expanded coverage.

We first validate pass@$k$ in the setting of classifier-free guidance (CFG)~\citep{ho2022classifier}, where the tradeoff between stronger single-sample quality and reduced diversity is well established. This provides a natural sanity check: if pass@$k$ reflects distribution coverage, it should recover the known contraction induced by strong guidance. Indeed, high CFG improves pass@$1$, but its advantage shrinks and eventually reverses at large $k$.

With this validation in hand, we use pass@$k$ to study few-step distillation. Across several widely used model families, distilled students often outperform their teachers at small $k$ but fall behind at large $k$, revealing a previously obscured loss of coverage. A controlled comparison under a shared teacher and evaluation protocol further traces this behavior to the distillation objective: distribution-matching losses concentrate probability on a narrower set of outputs, improving early-hit performance at the expense of large-budget coverage, whereas consistency- and trajectory-based losses better preserve the teacher's coverage. A complementary CLIP-space analysis shows that this contraction is accompanied by reduced within-prompt semantic diversity. Together, these results identify the distillation objective as a key factor in whether a few-step model preserves its teacher's distribution coverage or trades it for stronger single-draw performance.

We finally ask whether this finding extends beyond image generation. In few-step causal video generation~\citep{huang2026self,zhao2026causal}, we observe the same objective-dependent pattern: distribution-matching methods exhibit diminishing gains as the sampling budget grows, while consistency-style alternatives better preserve coverage.

Our contributions can be summarized as follows: (1) we introduce pass@$k$ as a diagnostic of distribution coverage for diffusion models, validating it on CFG where the coverage--quality tradeoff is independently known; (2) we identify an objective-level divide in diffusion distillation, showing that distribution-matching losses systematically contract coverage while consistency-based losses preserve it; and (3) we demonstrate that this divide holds across both image and video generation.

\section{Related Work}
\label{SEC:RELATED}

\paragraph{Text-to-image benchmark evaluation.}

Recent text-to-image benchmarks evaluate increasingly fine-grained aspects of prompt following. GenEval~\citep{ghosh2023geneval} emphasizes object-centric compositional correctness, GenEval2~\citep{kamath2025geneval} provides softer and more fine-grained atomic evaluation, DPG-Bench~\citep{hu2024ella} targets dense prompt following with VQA-style checks, GenAI-Bench~\citep{li2024genai} evaluates semantic text-to-image generation across diverse instruction categories, and T2I-CompBench~\citep{huang2023t2i} evaluates disentangled compositional skills such as attribute binding, spatial relations, and counting. These benchmarks are usually reported as single-sample or aggregated scores, which retain only the mean of the prompt-wise success probabilities. We use pass@$k$ on top of these same verifiers to reveal how that success probability is distributed across prompts as the sampling budget grows.

\paragraph{Distribution coverage evaluation.}

Prior work has addressed the quality--coverage tradeoff through diversity metrics such as the Vendi score~\citep{friedman2022vendi}, precision--recall decompositions for generative models~\citep{kynkaanniemi2019improved}, and distributional distances such as FID~\citep{heusel2017gans}. These metrics provide useful ImageNet~\citep{deng2009imagenet}-level diagnostics of realism and diversity, but they are not directly tied to whether a model can solve a particular text prompt under a benchmark verifier; moreover, representation-based distributional metrics can be directly optimized during training~\citep{yang2026representation}, limiting their reliability as independent evaluators. Our approach instead builds on prompt-level verifiers to measure coverage in terms of benchmark success, complemented by CLIP-space analysis of within-prompt semantic diversity.

\section{\texorpdfstring{Pass@$k$}{Pass@k} as a Lens for Distribution Coverage}
\label{SEC:METHOD}

\paragraph{Pass@$k$ on Text-to-image Generation Benchmarks.}

We borrow the pass@$k$ metric from code generation~\citep{chen2021evaluating}, where it measures the probability that at least one of $k$ independently sampled completions passes a verifier. To extend it to text-to-image generation, we treat each benchmark instance as a problem and the benchmark's automatic evaluator---e.g., the object, attribute, and spatial checks of GenEval~\citep{ghosh2023geneval}, the DSG-style VQA questions of DPG-Bench~\citep{hu2024ella}, or the semantic instruction-following evaluations of GenAI-Bench~\citep{li2024genai}---as the scoring oracle. A benchmark instance may be a single prompt or a prompt--category pair, depending on the benchmark's native protocol; in all cases, one sample denotes one independently generated image or image set scored by the corresponding evaluator.

For each benchmark instance, we draw $n \geq k$ independent samples from the model, let $c$ denote the number judged correct by the evaluator, and let $q_\theta(x)$ denote the population one-draw success probability of instance $x$. We report the unbiased estimator of \citep{chen2021evaluating} together with its population form:
\begin{equation}
  \operatorname{pass}@k
  =
  \mathbb{E}_{\text{instances}}
  \left[
  1 - \binom{n-c}{k}\Big/\binom{n}{k}
  \right],
  \qquad
  \operatorname{Pass}_\theta(k)
  =
  \mathbb{E}_{x}\!\left[1-(1-q_\theta(x))^k\right].
  \label{eq:passk}
\end{equation}
The estimator equals the probability that a uniformly random size-$k$ subset of the $n$ samples contains at least one success, averaged over instances, and $1-\operatorname{Pass}_\theta(k)$ is the $k$th moment of the prompt-wise failure probability $1-q_\theta(x)$: pass@$1$ retains only the mean success probability, whereas the pass@$k$ curve provides a sequence of nonlinear summaries of how success probability is distributed across benchmark instances.

For continuous evaluators, best-of-$k$ measures the expected maximum score among
$k$ independent generations. Given $n\geq k$ samples per prompt, we estimate it
by averaging the maximum over all uniformly chosen size-$k$ subsets, computed
exactly from the ascending order statistics $s_{p,(1)}\leq\cdots\leq s_{p,(n)}$:
\begin{equation}
  \widehat B(k)=\frac{1}{|\mathcal P|}\sum_{p\in\mathcal P}
  \sum_{j=k}^{n}\frac{\binom{j-1}{k-1}}{\binom nk}s_{p,(j)}.
  \label{eq:bestofk}
\end{equation}
The weight is the probability that the $j$th ordered sample is the maximum of
the subset. For a fixed pool of independent, identically distributed samples,
this estimator is unbiased and reduces to Eq.~\ref{eq:passk} for binary scores.
It probes access to the upper tail of continuous evaluator scores as the budget grows.

\paragraph{Experiment Setup.}

We use pass@$k$ and best-of-$k$ as unified diagnostics of distribution coverage. For each model and experimental condition, we generate $n$ independent samples per prompt using the same prompt set, resolution, benchmark split, and evaluator. Unless otherwise noted, all models are sampled with their default \texttt{diffusers}~\citep{vonplaten2022diffusers} inference pipelines and hyperparameters. We then compute the appropriate statistic for $k \in \{1,2,4,\ldots,n\}$, producing a coverage profile rather than a single score. The $k{=}1$ endpoint recovers standard single-sample performance, while the growth of the curve measures how much additional success becomes accessible at larger $k$.

All comparisons hold the sampling budget fixed across methods and differ only in the intervention being studied: the guidance scale for CFG experiments, or the student checkpoint for distillation experiments. The curves distinguish three budget-dependent behaviors. An improvement across the measured curve indicates greater access to successful generations at those budgets. A gain at $k{=}1$ that saturates or reverses indicates that the one-draw advantage does not persist with additional samples. Conversely, weaker one-draw performance followed by stronger large-$k$ performance indicates lower early-hit reliability but greater measured benefit from additional sampling. Because automatic evaluators have nonzero false-positive rates that accumulate at large budgets, we base claims on crossovers at moderate budgets and verify in Appendix~\ref{app:fp} that all reported comparisons survive an FP-immune variant of pass@$k$ and a direct calibration of evaluator false-positive rates.

\section{\texorpdfstring{Pass@$k$}{Pass@k} Profiles Distribution Coverage under CFG}
\label{SEC:CFG}

\paragraph{Background.} Classifier-free guidance~\citep{ho2022classifier} is one of the most widely used test-time techniques for improving prompt alignment in text-to-image diffusion models. During sampling, CFG combines the conditional and unconditional model predictions to amplify the direction associated with the text condition. In a common noise-prediction parameterization, the guided prediction is written as:
\begin{equation}
    \epsilon_{\mathrm{cfg}}(x_t, c)
    =
    \epsilon_{\theta}(x_t, \varnothing)
    +
    w \left(
        \epsilon_{\theta}(x_t, c)
        -
        \epsilon_{\theta}(x_t, \varnothing)
    \right),
\end{equation}
where $c$ is the text prompt, $\varnothing$ denotes the empty condition, and $w$ is the guidance scale.

\begin{figure}[tbp]
\centering
\includegraphics[width=0.95\linewidth]{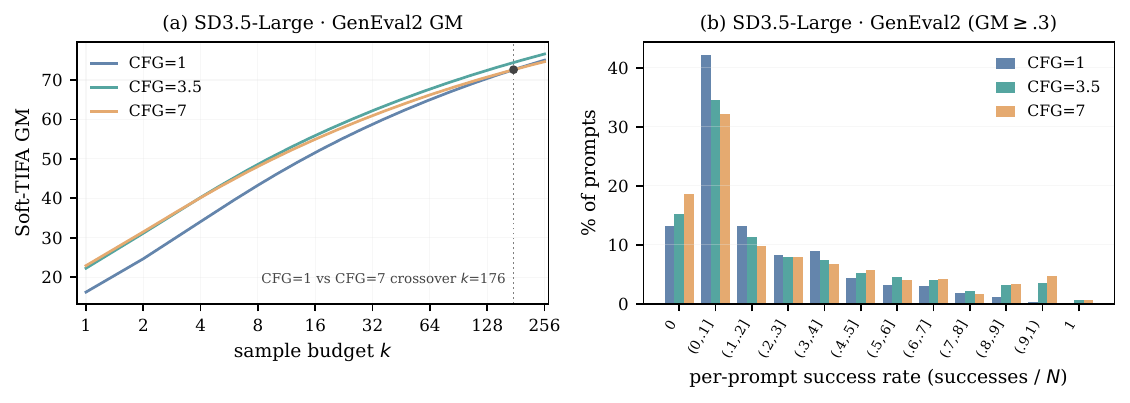}
\caption{\textbf{SD3.5-Large CFG sweep experiment on GenEval2.} (a) Best-of-$k$ GenEval2 Soft-TIFA GM for CFG$\,\in\{1,3.5,7\}$. The dotted line marks where CFG$=1$ overtakes CFG$=7$. (b) Distribution of per-prompt success rates, where success is Soft-TIFA GM$\geq0.3$ over $N{=}256$ samples---a lenient cutoff, shared by all GenEval2 per-prompt diagnostics, that credits any mostly-correct sample (Appendix~\ref{sec:distributions}).}
\label{fig:cfg-collapse}
\end{figure}

Larger guidance scales often improve prompt adherence and visual fidelity at the single-sample level. However, CFG also changes the sampling distribution by reweighting the reverse process toward samples that are more strongly favored by the condition. This motivates a coverage question: does guidance improve success broadly across prompts, or is its advantage concentrated at small sampling budgets?

\paragraph{Experimental Analysis.}

Fig.~\ref{fig:cfg-collapse} presents results for SD3.5-Large on GenEval2. GenEval2 decomposes each prompt into atomic requirements and assigns each a VQA-based soft score; Soft-TIFA GM is the geometric mean of these scores and serves as the prompt-level correctness measure~\citep{kamath2025geneval}. Panel~(a) shows that CFG$=7$ achieves the strongest best-of-$1$ score, but its advantage erodes as additional samples are drawn. CFG$=3.5$ overtakes it at a small budget and remains strongest thereafter, while CFG$=1$ eventually overtakes CFG$=7$. Panel~(b) explains this crossover through the distribution of per-prompt success rates. As guidance increases, probability mass moves away from prompts with low-but-nonzero success rates and toward both extremes: more prompts are solved reliably, but more remain unsolved throughout the measured sample pool. Together, these results provide evidence of prompt-level coverage contraction under high CFG---stronger guidance improves early-hit success but yields smaller marginal gains from additional sampling---which we term benchmark-conditioned mode collapse. Two caveats delimit this claim: CFG serves here as a controlled knob for demonstrating that pass@$k$ separates behaviors indistinguishable at $k{=}1$, not as evidence that guidance is harmful; and since extreme guidance also introduces perceptual artifacts that a VQA evaluator may reject, the curves measure benchmark-conditioned coverage, the quantity a verifier-based best-of-$k$ pipeline consumes, with mechanism-level diversity analyzed directly in Section~\ref{SEC:DISTILLATION}. The complete CFG sweep and corresponding per-prompt distributions are provided in Appendix~\ref{app:cfg_full}.

\section{Distribution Coverage in Few-Step Distillation}
\label{SEC:DISTILLATION}

\begin{figure}[!t]
\centering
\includegraphics[width=\linewidth]{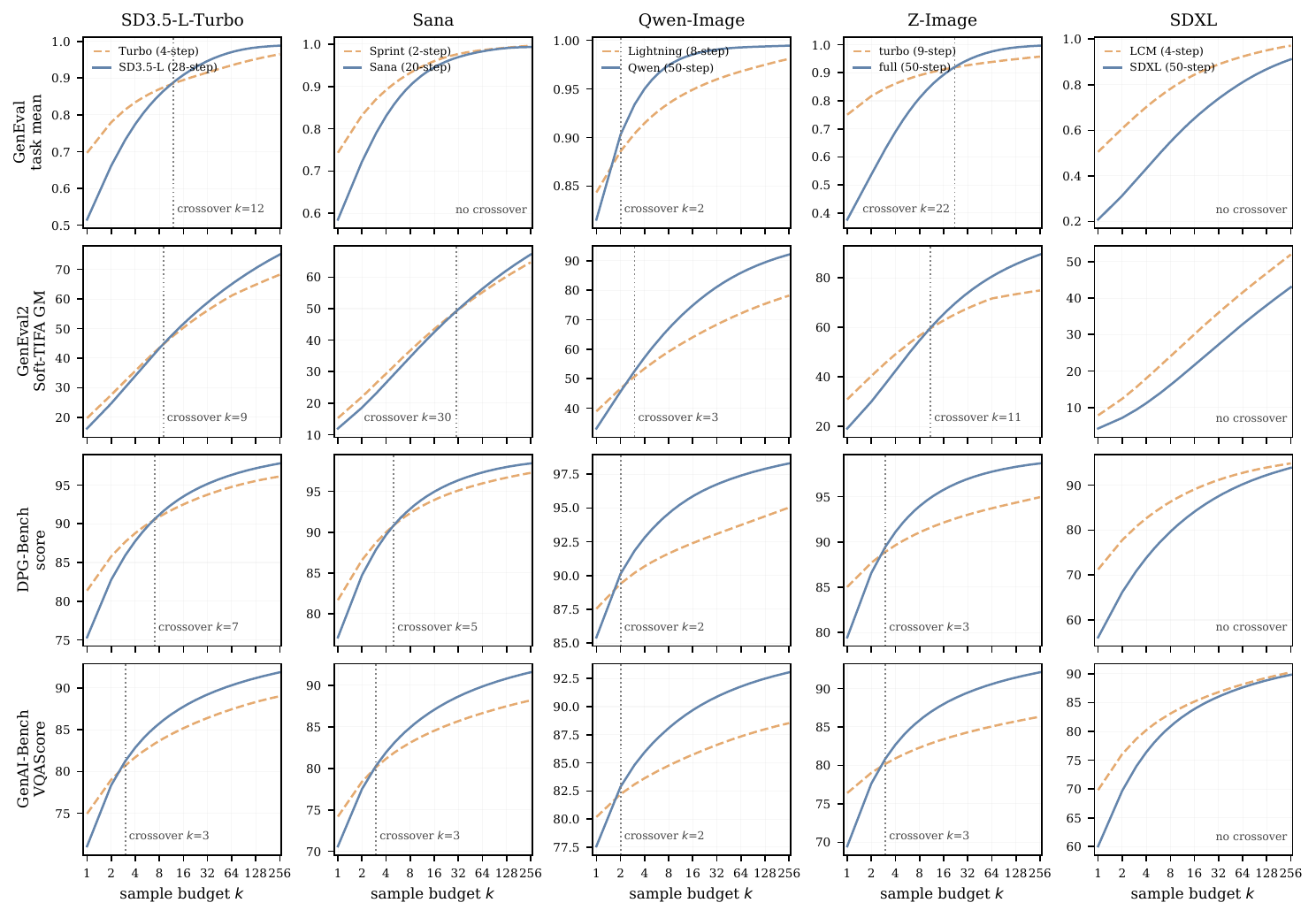}
\caption{\textbf{Teacher--student performance across sampling budgets.}
Pass@$k$ on GenEval and expected best-of-$k$ on GenEval2, DPG-Bench, and
GenAI-Bench for five distillation families ($N{=}256$ per prompt). Solid and
dashed curves denote the teacher and student, respectively; dotted lines mark
the first teacher--student crossover.}
\label{fig:distill-matrix}
\end{figure}

\paragraph{Background.}
Diffusion distillation compresses the iterative reverse process of a multi-step
teacher into a student that generates in one or a few denoising steps. Recent
methods have substantially narrowed the fidelity gap to their teachers, making
one-step synthesis increasingly practical~\citep{yin2024improved,chen2025sana,wang2026uni,geng2026mean,song2023consistency}.
Diffusion distillation can be divided into two broad
paradigms base on training objectives: distribution matching and consistency distillation. Practical
recipes may combine either principle with adversarial or regression
losses~\citep{chen2025sana,yin2024improved}.
For example, Sana-Sprint combines consistency and latent adversarial distillation.

\paragraph{Distribution matching.}
Let $p_{\theta}(x\mid c)$ and $p_{\phi}(x\mid c)$ denote the conditional output
distributions of the teacher and student. Distribution-matching distillation
directly reduces a discrepancy between them, commonly expressed as
\begin{equation}
    \mathcal{L}_{\mathrm{DM}}(\phi)
    =
    \mathbb{E}_{c}\!\left[
        D_{\mathrm{KL}}\!\left(
            p_{\phi}(\cdot\mid c)
            \,\|\,
            p_{\theta}(\cdot\mid c)
        \right)
    \right].
    \label{eq:distribution-matching}
\end{equation}
Because the output densities are implicit, this gradient is estimated using
teacher and student score functions and is often supplemented by adversarial
or regression losses~\citep{yin2024improved,sauer2024adversarial}. The student
is thus trained to match the teacher distribution without reproducing a
specific teacher sampling trajectory.

\paragraph{Consistency distillation.}
Consistency methods instead require the student to make the same endpoint
prediction from different states on a teacher probability-flow ODE
trajectory~\citep{luo2023latent,chen2025sana,geng2026mean}. Let $\hat{x}_{s}$ be obtained by
advancing $x_t$ from time $t$ to $s<t$ with the teacher solver, and let
$f_{\phi}$ map a noisy state to the trajectory endpoint. A standard objective
has the form
\begin{equation}
    \mathcal{L}_{\mathrm{CD}}(\phi,\bar{\phi})
    =
    \mathbb{E}\!\left[
        \lambda(t)\,
        d\!\left(
            f_{\phi}(x_t,t,c),
            \operatorname{sg}\!\left[
                f_{\bar{\phi}}(\hat{x}_{s},s,c)
            \right]
        \right)
    \right],
    \label{eq:consistency-distillation}
\end{equation}
where $\bar{\phi}$ is an exponential-moving-average target,
$\operatorname{sg}$ stops gradients, and $d$ is a regression distance. Once
trained, the endpoint map can be evaluated once or composed for a small number
of steps.
We group our teacher-supervised MeanFlow variant under consistency-style
distillation because its average-velocity identity corresponds to a
differential consistency condition on the induced flow
map~\citep{geng2026mean}.

Although one- and few-step students now produce sharp, high-scoring samples,
visual fidelity does not guarantee that they preserve the teacher's
conditional distribution. A student can concentrate probability on fewer
valid outcomes, yielding more homogeneous samples even when its one-sample
score improves. Compared with fidelity and sampling speed, this loss of
prompt-conditioned diversity has received much less systematic evaluation.
Moreover, it cannot be resolved from FID or a one-sample benchmark score alone~\citep{ravishankar2026setting}.
We therefore ask whether the student preserves the teacher's distribution
coverage as the sampling budget increases.

\subsection{Coverage Contraction and Conditional Mode Collapse}

Fig.~\ref{fig:distill-matrix} presents five distilled students with their corresponding multi-step baselines: SD3.5-Large-Turbo~\citep{sauer2024adversarial} with SD3.5-Large~\citep{esser2024scaling}, Sana-Sprint~\citep{chen2025sana} with Sana~\citep{xie2024sana}, Qwen-Image-Lightning with
Qwen-Image~\citep{wu2025qwen}, LCM-SDXL~\citep{luo2023latent} with
SDXL~\citep{rombach2022high}, and Z-Image-Turbo with
Z-Image~\citep{cai2025z}. All evaluations use $N{=}256$ samples per prompt. We
report expected best-of-$k$ on GenEval2, DPG-Bench, and GenAI-Bench, and
pass@$k$ on GenEval. Every student leads its teacher at $k{=}1$, but the teacher overtakes within budget in 16 of the 20 comparisons.
Appendix~\ref{sec:explore} breaks these crossovers down by benchmark subscore
and provides prompt-level coverage diagnostics.

\begin{figure}[t]
\centering
\includegraphics[width=0.9\linewidth]{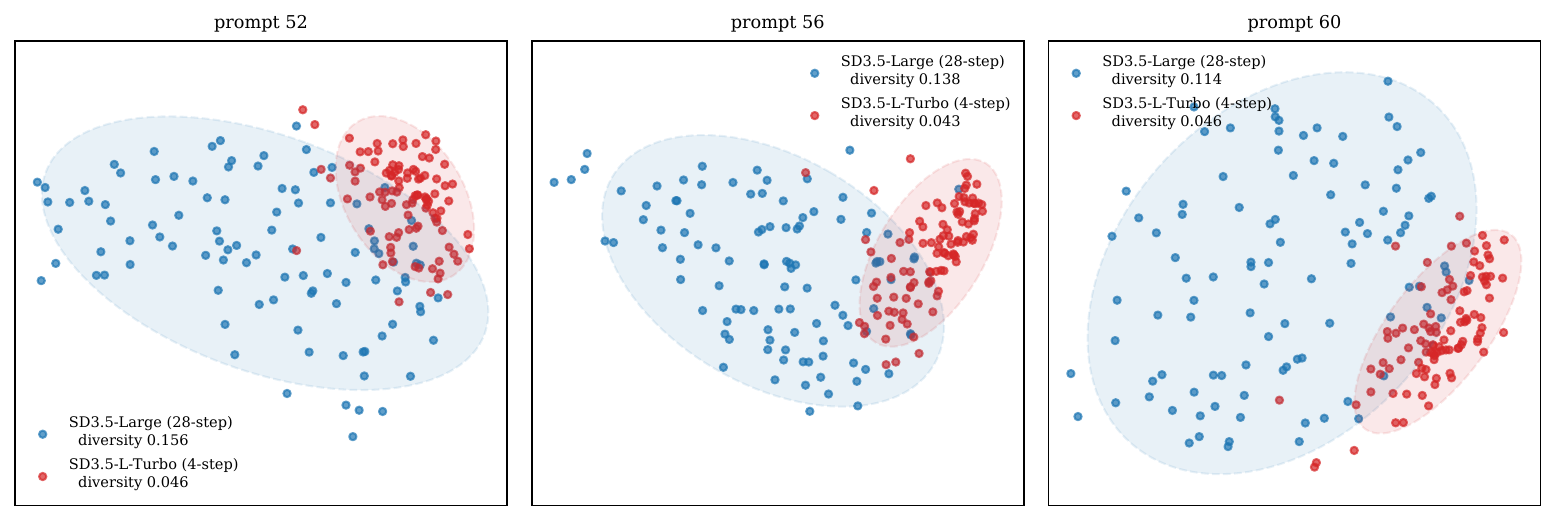}
\caption{\textbf{Distilled samples are more concentrated on matched-correct
prompts.} Joint top-2 PCA projections of CLIP embeddings from SD3.5-Large and
SD3.5-Large-Turbo on three matched-correct prompts ($100$ samples per model,
per-model $2\sigma$ ellipses). Each panel reports the mean pairwise cosine
distance in the original CLIP space.}
\label{fig:fx-embed}
\end{figure}

The pass@$k$ crossover shows that the distilled student's early advantage does
not persist as the sampling budget grows. We next ask whether this contraction
is accompanied by reduced diversity among generations for the same prompt.
We illustrate this with SD3.5-Large and its Turbo student.
Fig.~\ref{fig:fx-embed} compares $100$ samples from each model on three
GenEval prompts that both solve perfectly, using CLIP ViT-L/14
embeddings~\citep{radford2021learning}. Despite identical correctness, the
teacher's mean pairwise CLIP distance is $2.5$--$3.4$ times larger, and the PCA
plots show the Turbo samples clustering much more tightly.

LCM-SDXL\citep{luo2023latent} is the only student that remains ahead of its teacher across all four
benchmarks. Its consistency objective learns to traverse the teacher's
denoising trajectory in fewer steps, and the sustained gains show that this
trajectory-shortening strategy can preserve coverage rather than inevitably
induce mode collapse. Sana-Sprint, however, combines continuous-time consistency
distillation (sCM) with latent adversarial distillation
(LADD)~\citep{chen2025sana}. Its teacher--student crossovers therefore reflect
the behavior of a hybrid training objective and cannot be attributed to
consistency distillation alone. Section~\ref{SEC:LOSS} next examines why coverage behavior
differs across training objectives in a controlled FLUX setting.

\subsection{What in Distillation Causes the Collapse? A Controlled Loss Ablation}
\label{SEC:LOSS}

\begin{figure}[!t]
\centering
\includegraphics[width=0.9\linewidth]{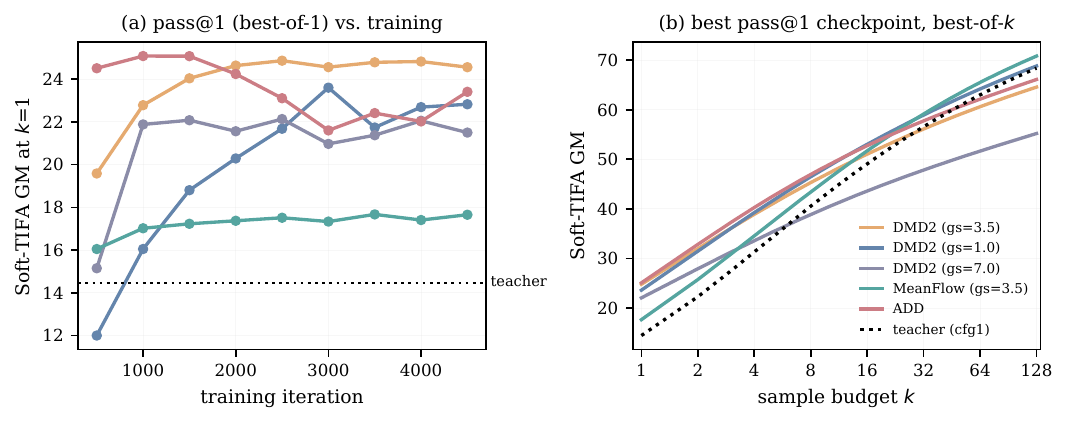}
\caption{
\textbf{Distillation objectives shape coverage across training and sampling budgets.}
\emph{(a)} GenEval2 GM at $k{=}1$ across training checkpoints.
\emph{(b)} Best-of-$k$ curves for the checkpoint from each run with the strongest
$k{=}1$ performance.}
\label{fig:2x2}
\end{figure}

\begin{figure}[!t]
\centering
\includegraphics[width=0.9\linewidth]{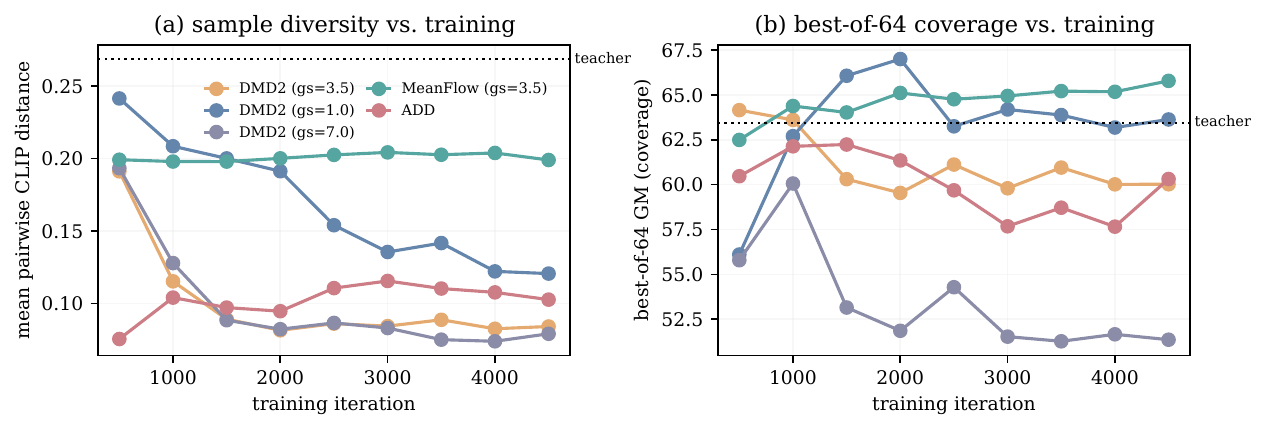}
\caption{
\textbf{Coverage degradation is accompanied by diversity collapse.}
Along the same training trajectories, \emph{(a)} reports mean pairwise CLIP
distance within each prompt and \emph{(b)} reports GenEval2 GM best-of-$64$.
DMD2 contracts in representation space as its coverage declines, whereas
MeanFlow preserves diversity while improving coverage.}
\label{fig:collapse}
\end{figure}

To understand why some students collapse more severely than others, we group
distillation methods by their training objectives. Distribution-matching
methods such as DMD2 optimize the final-sample distribution and often include
adversarial or reverse-KL-like terms that are known to be mode-seeking~\citep{wang2026uni}.
Consistency- and trajectory-based methods such as MeanFlow instead compress
the teacher's sampling path by regressing its average velocity field. Although Section~\ref{SEC:DISTILLATION} shows that
few-step distillation generally reduces large-$k$ coverage, those comparisons
also change the teacher, architecture, and training data. LCM-SDXL is a notable
exception in Fig.~\ref{fig:distill-matrix}, but it belongs to a different model
family. We therefore use a controlled FLUX.1-dev study to isolate the role of
the distillation objective.

\paragraph{Controlled design.}
We distill the same FLUX.1-dev teacher into the same four-step student
architecture on GPIC 1M~\citep{chandrasegaran2026gpic} at $512\times512$
resolution, varying the distillation loss:
\begin{itemize}
\item \textbf{DMD2}~\citep{yin2024improved} combines distribution matching with
an adversarial term against real data, a recipe adopted by several recent
large text-to-image models~\citep{wu2025qwen,zhao2026qwen,cai2025z};
\item \textbf{ADD}~\citep{sauer2024adversarial} uses an adversarial objective
against real data without the DMD distribution-matching term;
\item \textbf{MeanFlow}~\citep{geng2026mean} regresses the teacher's average
velocity field without an adversarial or distribution-matching term.
\end{itemize}
We evaluate each training checkpoint with $128$ GenEval2 samples per prompt. All students use an
embedded guidance scale of $3.5$; for DMD2, we additionally evaluate scales
$1.0$ and $7.0$. The FLUX.1-dev teacher uses a 50-step Euler sampler without
CFG.

\paragraph{Result 1: distribution-matching and adversarial objectives exhibit coverage collapse.}
Fig.~\ref{fig:2x2} reports $k{=}1$ quality across training and the best-of-$k$
curves of each run's strongest $k{=}1$ checkpoint.
Fig.~\ref{fig:collapse}(b) tracks best-of-$64$ GM across training. DMD2 follows
the same non-monotonic trajectory at each guidance scale: coverage improves
early and then declines. ADD also peaks early and remains below the teacher,
whereas MeanFlow improves throughout training and finishes above it. Higher
guidance further amplifies the late-training decline within DMD2, but the contrast
between DMD2 and MeanFlow remains under their shared standard guidance.

\paragraph{Separating objective and guidance effects.}
We use two complementary comparisons to distinguish the role of the
distillation objective from that of guidance. First, at the same embedded
guidance scale of $3.5$, DMD2 and MeanFlow exhibit different training dynamics:
DMD2 loses coverage and diversity during later training, whereas MeanFlow
preserves diversity while improving coverage. These contrasting trajectories
suggest that the distillation objective affects coverage and diversity even
at a fixed guidance scale. Second, holding the DMD2
objective fixed, we vary guidance over $\{1.0, 3.5, 7.0\}$. Even at guidance
$1.0$, best-of-$64$ performance declines after its peak alongside decreasing
CLIP diversity, while stronger guidance amplifies the late-training
degradation. The guidance ablation further suggests that elevated guidance
amplifies, but does not fully explain, the observed collapse.

\paragraph{Result 2: pass@$k$ degradation is accompanied by reduced diversity.}
Fig.~\ref{fig:collapse} measures within-prompt diversity along the same training
trajectories. MeanFlow maintains nearly constant CLIP diversity while its
coverage improves. DMD2 instead contracts sharply in representation space,
and its later checkpoints remain substantially below MeanFlow and the teacher.
The decline in DMD2's pass@$k$ performance therefore coincides with a loss of
generative diversity rather than only a change in the benchmark score.
Although ADD's CLIP diversity remains relatively stable across training
checkpoints, it stays substantially below the teacher's, indicating a
persistent diversity deficit rather than progressive contraction during the
observed training period.

\paragraph{Mechanism and scope.}
The results follow the training objective. Distribution-matching and
adversarial objectives place pressure on the student to concentrate probability
mass in high-density regions of the target distribution. MeanFlow instead
learns a shortened teacher trajectory without directly reshaping its output
distribution. In this controlled setting, the divergence-based objectives are
associated with reduced diversity and coverage collapse, whereas the
consistency-style objective preserves both. Final-checkpoint curves and per-prompt
success profiles are provided in Appendix~\ref{app:passk_sweeps}.

\section{Distribution Coverage in Video Forcing Models}
\label{app:video_forcing_collapse}

\begin{figure}[!t]
\centering
\includegraphics[width=0.9\linewidth]{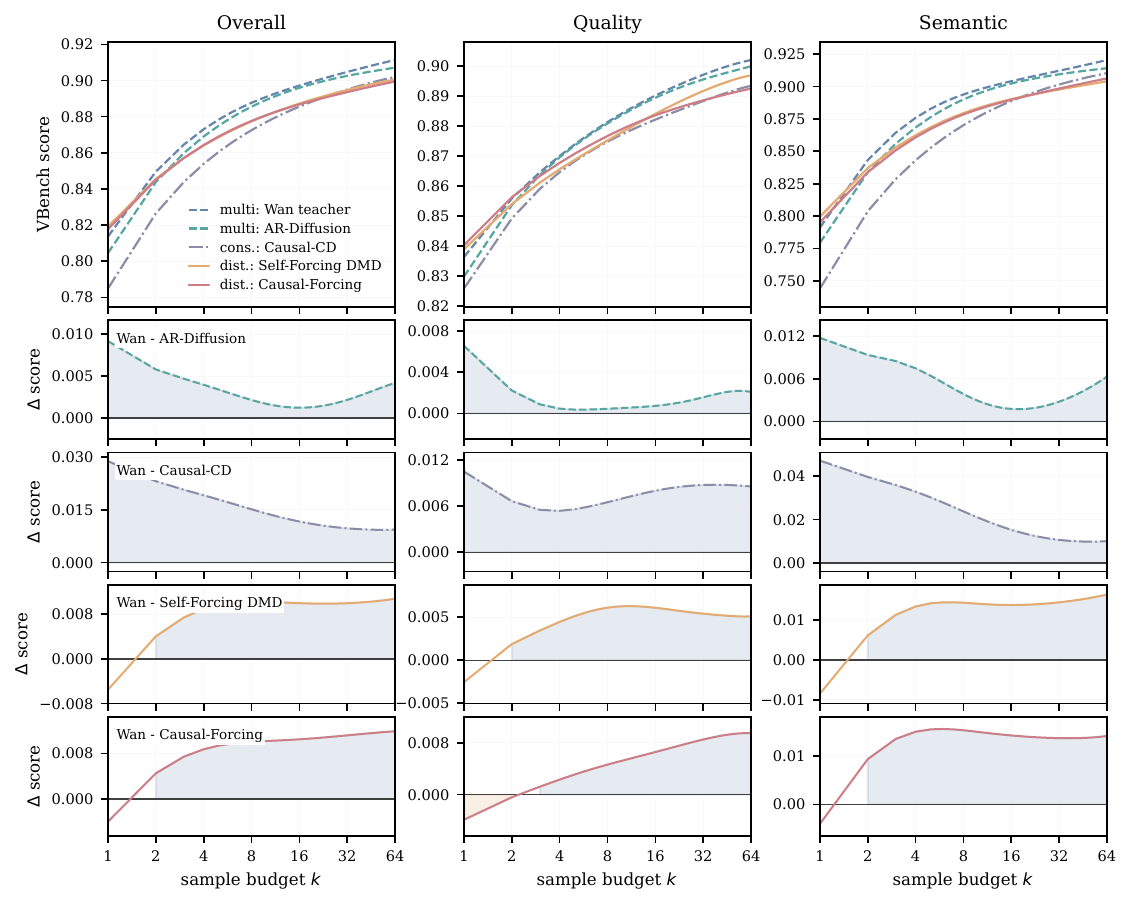}
\caption{\textbf{VBench-20 best-of-$k$ curves for video forcing models.} Each curve uses the exact subset-averaged estimator in Eq.~\ref{eq:bestofk}. Solid curves denote distribution-matching methods (Self-Forcing DMD and Causal-Forcing), dash-dot curves denote
consistency distillation (Causal-CD), and dashed curves denote multi-step
baselines. The four lower rows show
$\Delta_m(k)=s_{\mathrm{Wan}}(k)-s_m(k)$ for AR-Diffusion, Causal-CD,
Self-Forcing DMD, and Causal-Forcing, respectively. Positive differences favor Wan.}
\label{fig:vbench20-video-passk}
\end{figure}

\paragraph{Video forcing methods.}

With the insight from text-to-image diffusion distillation, we extend our
evaluation to few-step video diffusion models. Video forcing methods accelerate
autoregressive video generation by distilling a multi-step video diffusion
process into a shorter sampler, while preserving the causal rollout structure
needed for long video generation. We evaluate two representative families:
self-forcing~\citep{huang2026self} and causal forcing~\citep{zhao2026causal}, compared with their teacher Wan model~\citep{wan2025wan}. Self-forcing is a pure
distribution-matching distillation approach. Causal
forcing includes two kinds of few-step approaches, allowing us to separate a
distribution-matching variant from a consistency-distillation variant.

\paragraph{Benchmark selection.} We evaluate VBench-20 using a fixed set of
20 prompts for each of 16 VBench dimensions and $n=64$ independently sampled
videos per dimension prompt. For each dimension $d$, we apply
Eq.~\ref{eq:bestofk} to the normalized per-video scores $s_{p,i,d}$ and average
over its prompt set $\mathcal P_d$. Thus each point is the exact expected
maximum over uniformly chosen size-$k$ subsets of the 64 videos; $k=1$ uses
the mean of all 64 scores. We retain the source evaluation's zero scores for
missing color records (7--16 of 1,280 records per method).
We report every integer $k\in[1,64]$ and three aggregates: quality averages
seven dimension curves with weight $0.5$ for dynamic degree and $1$ otherwise;
semantic equally averages nine dimension curves; overall equally averages
quality and semantic. Selection occurs separately within each dimension.

\paragraph{Analysis.}

Both the quality and semantic best-of-$k$ curves in
Fig.~\ref{fig:vbench20-video-passk} show signs of mode collapse in the
distribution-matching models. Self-Forcing DMD and Causal-Forcing initially
outperform Wan, but their advantage quickly reverses as the sampling budget
grows, consistent with our controlled image study in Section~\ref{SEC:LOSS}.

Multi-step AR-Diffusion and trajectory-based consistency distillation show a different pattern. Their quality scores remain below the
Wan teacher across the evaluated budgets, but their semantic scores tend to
catch up as more samples are drawn. This suggests that semantic diversity
may be better preserved in these approaches, even when visual quality lags
behind. These results motivate further work on distillation methods that
improve visual quality while preserving semantic diversity.

\let\FloatBarrier\relax   %
\section{Conclusion}
\label{SEC:CONCLUSION}

We use pass@$k$ to probe whether few-step diffusion students truly inherit their teachers' distribution coverage. After validating the diagnostic on CFG, where the coverage--quality tradeoff is independently known, we show that the answer depends on the distillation objective: distribution-matching and adversarial losses concentrate the student's output distribution, improving single-draw performance at the cost of large-budget coverage, whereas consistency and trajectory-based losses better preserve the teacher's coverage. The same divide holds in few-step causal video generation, suggesting it reflects a fundamental property of the training objective rather than a domain-specific artifact. These results challenge the common assumption that distilled models are drop-in replacements for their teachers, and highlight that the choice of distillation loss has consequences for distribution coverage that single-draw metrics do not reveal.

\paragraph{Limitations.}
Pass@$k$ is deliberately benchmark-relative: it characterizes access to
successful generations under a specified evaluation criterion rather than
every form of perceptual or semantic diversity. Its large sample pools also
limit how many models and training configurations can be evaluated,
particularly proprietary systems. We hope our findings motivate greater
attention to diversity-preserving distillation methods that retain both
sampling efficiency and distribution coverage.

\clearpage

\bibliography{iclr2027_conference}
\bibliographystyle{iclr2027_conference}

\clearpage
\appendix
\section{Additional CFG Results}
\label{app:cfg_full}

\subsection{Full Guidance Sweep}

\begin{figure}[!h]
\centering
\includegraphics[width=\linewidth]{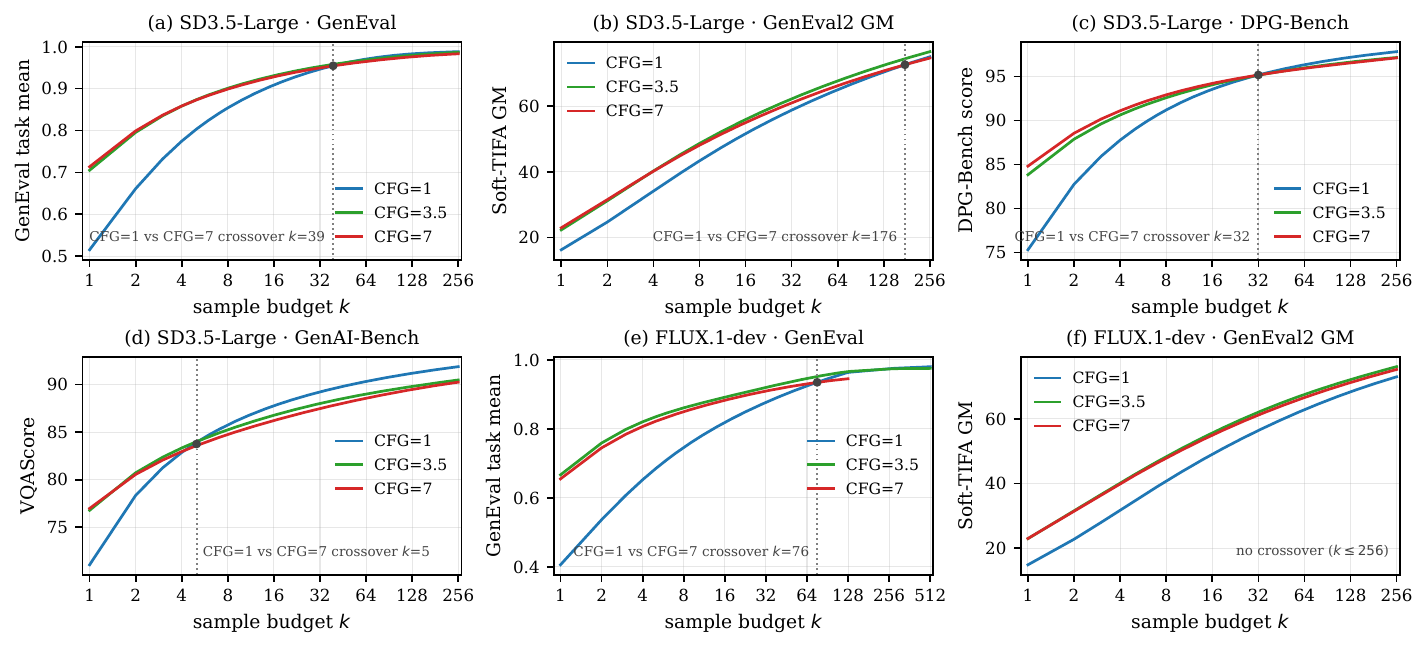}
\caption{\textbf{Full CFG sweep.} Pass@$k$ and best-of-$k$ curves for
SD3.5-Large (a--d) and FLUX.1-dev (e--f) under CFG$\in\{1,3.5,7\}$.
Dotted lines mark where CFG$=1$ first matches or overtakes CFG$=7$.}
\label{fig:cfg-full}
\end{figure}

Fig.~\ref{fig:cfg-full} extends the selected comparison in
Fig.~\ref{fig:cfg-collapse} to all evaluated model--benchmark pairs and to
CFG$\in\{1,3.5,7\}$. Stronger guidance
usually improves performance at small sampling budgets, but the advantage
narrows and often reverses as $k$ increases. The clearest reversal occurs on
GenAI-Bench (Fig.~\ref{fig:cfg-full}d). Moderate guidance is not uniformly
harmful: for FLUX.1-dev on GenEval2 (Fig.~\ref{fig:cfg-full}f), CFG$=3.5$
remains ahead throughout the measured range. This suggests that well-tuned
guidance can improve one-sample performance without the same loss in
large-budget gains observed under stronger guidance.

\subsection{Per-Prompt Success Rates}
\label{sec:distributions}

\begin{figure}[!h]
\centering
\includegraphics[width=\linewidth]{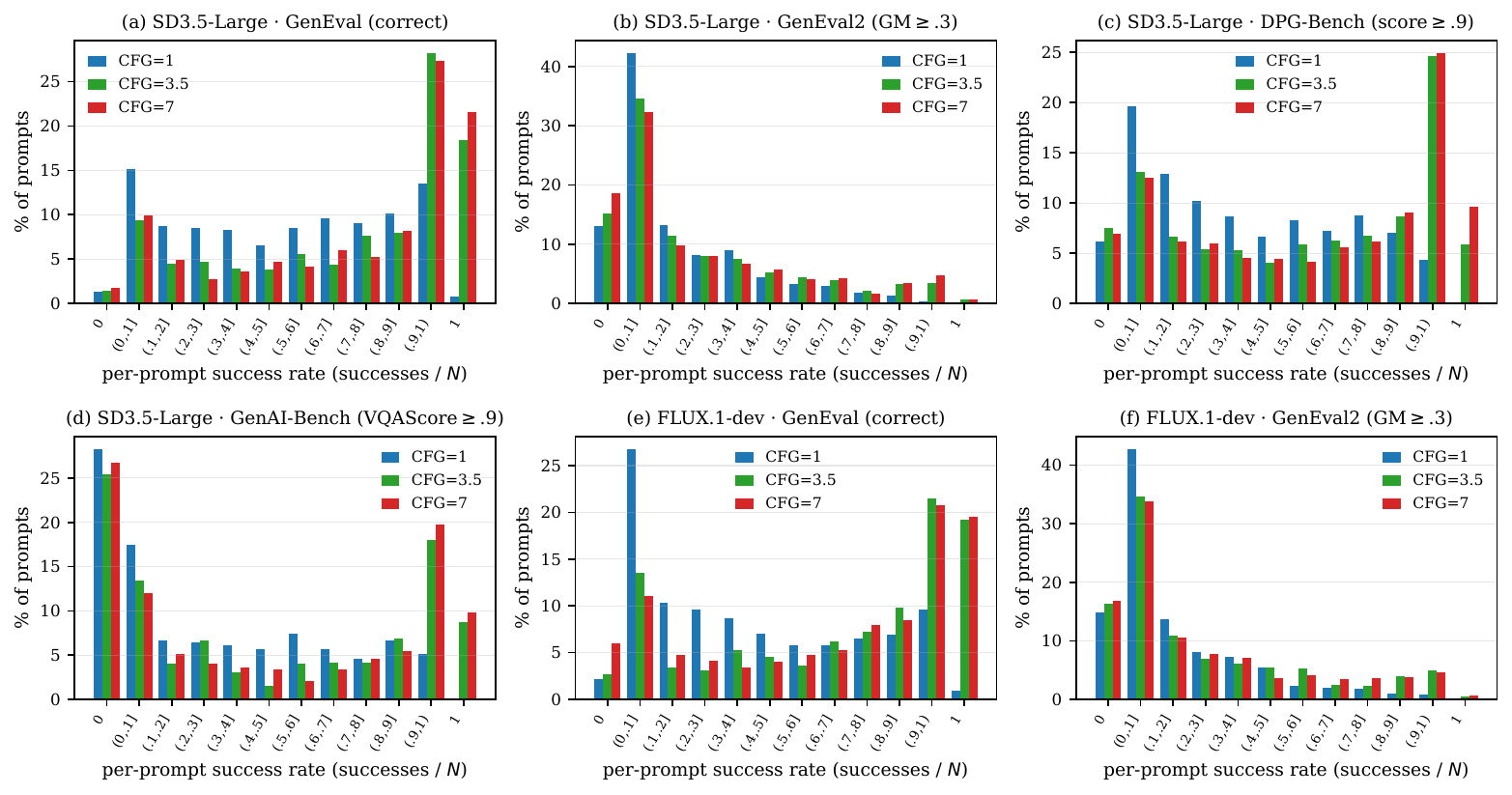}
\caption{\textbf{Per-prompt success-rate distributions for the CFG sweeps.}
Panel order follows Fig.~\ref{fig:cfg-full}.}
\label{fig:cfg-dist}
\end{figure}

Fig.~\ref{fig:cfg-dist} shows the per-prompt success rates underlying the
curves. A sample
counts as successful when it is correct on GenEval, reaches
Soft-TIFA GM$\geq0.3$ on GenEval2, reaches $0.9$ on DPG-Bench, or reaches
VQAScore$\geq0.9$ on GenAI-Bench. The GenEval2 cutoff is deliberately
lenient and is shared by every GenEval2 per-prompt diagnostic in the paper
(Figs.~\ref{fig:cfg-collapse} and~\ref{fig:2x2-dist}) as well as by the
pass@$k$ re-analysis of Appendix~\ref{app:fp}: a low cutoff asks whether the
model \emph{ever} produces a mostly-correct sample for a prompt, which is
the coverage notion that pass@$k$ targets, whereas a cutoff near $1$
increasingly reflects single-sample polish---precisely the quantity that
guidance improves---rather than coverage. No quantitative claim depends on
this binarization: best-of-$k$ curves are computed on the raw continuous
scores (Eq.~\ref{eq:bestofk}), and Appendix~\ref{app:fp} reports how the
crossovers shift under stricter cutoffs ($\tau=0.7$ and $0.9$).
Higher guidance shifts probability mass out
of the intermediate-success prompts and toward the extremes. This makes
already tractable prompts more reliable, but leaves fewer intermediate prompts
that can benefit from additional samples. Indeed, in five of the six settings
in Fig.~\ref{fig:cfg-dist}, the CFG$=1$ baseline leaves fewer prompts in the
zero-success bin than CFG$=7$, including SD3.5-Large on GenEval2 and DPG-Bench
and FLUX.1-dev on GenEval and GenEval2; stronger CFG therefore does not improve
generation uniformly across prompts, but concentrates probability on prompts
the model already solves and contracts the effective sampling distribution.

\section{Additional Distillation Results}
\label{sec:explore}

This section provides the complete crossover statistics and per-family
diagnostics behind the distillation results in the main paper.

\subsection{Crossovers Across Families}

\begin{figure}[!ht]
\centering
\includegraphics[width=\linewidth]{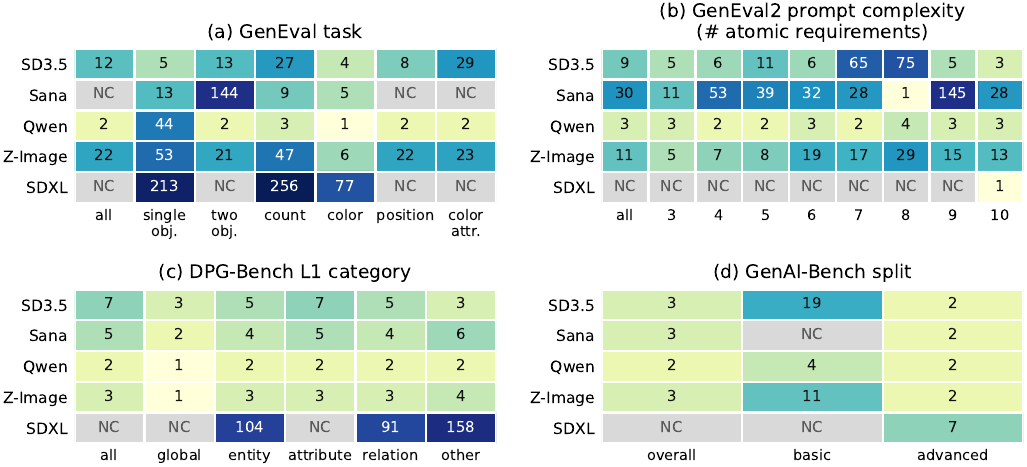}
\caption{\textbf{Teacher--student crossover budgets by benchmark subscore.}
Rows are model families and columns are benchmark-specific tasks, atom counts,
categories, or splits. Darker cells indicate later crossovers; gray cells
marked NC do not cross within $N{=}256$.}
\label{fig:subscore-crossovers}
\end{figure}

Fig.~\ref{fig:subscore-crossovers} breaks the aggregate crossovers in
Fig.~\ref{fig:distill-matrix} into benchmark-specific subscores. Each cell is
the smallest integer $k$ at which the teacher matches or exceeds the student;
$k{=}1$ means that the teacher already leads, and NC denotes no crossover
within $N{=}256$. Crossovers are widespread but family-dependent: they occur across most
subscores for SD3.5-Turbo, Qwen-Image-Lightning, and Z-Image-Turbo, whereas
LCM-SDXL remains ahead on every aggregate benchmark and is overtaken only on
selected subscores, consistent with better coverage preservation under
consistency distillation.

\FloatBarrier

\subsection{Prompt-Level Coverage Diagnostics}

\paragraph{Success redistribution across prompts.}

\begin{figure}[!h]
\centering
\includegraphics[width=\linewidth]{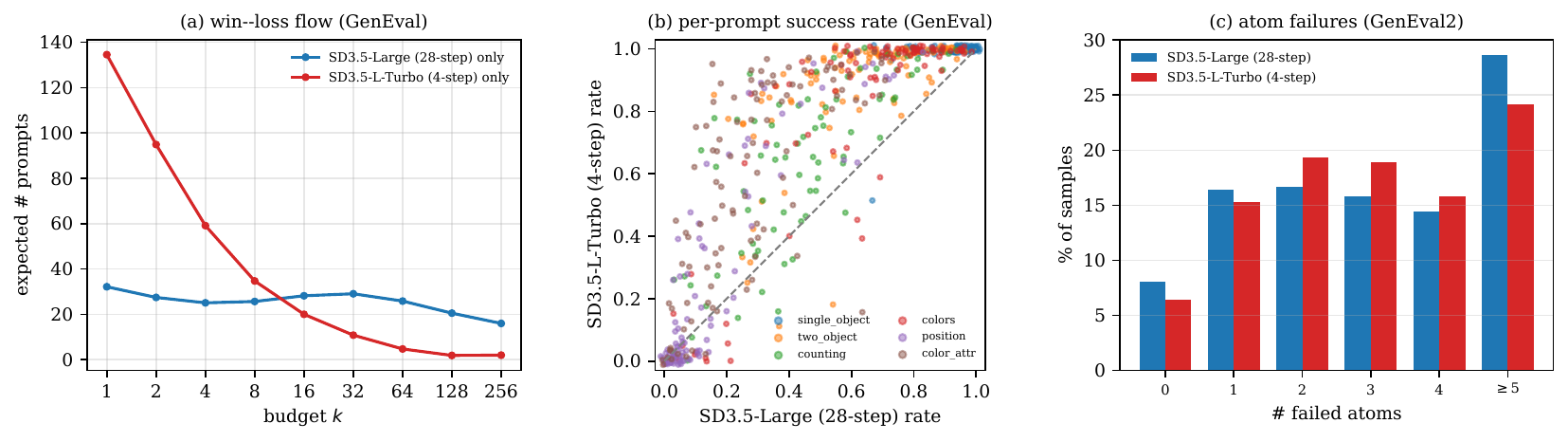}
\caption{\textbf{Per-prompt diagnostics for SD3.5-Large and Turbo.}
Panels show exclusive wins over $k$, per-prompt success rates, and GenEval2
failed-atom counts.}
\label{fig:fxd-fam-turbo}
\end{figure}

\begin{figure}[!h]
\centering
\includegraphics[width=\linewidth]{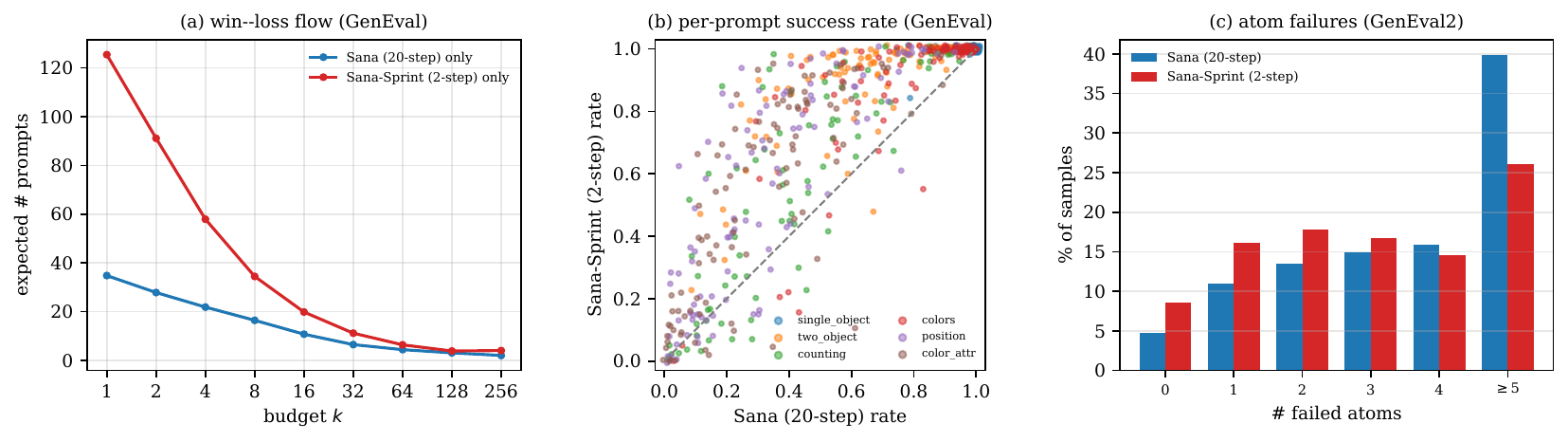}
\caption{\textbf{Per-prompt diagnostics for Sana and Sana-Sprint.}
Panels follow Fig.~\ref{fig:fxd-fam-turbo}.}
\label{fig:fxd-fam-sana}
\end{figure}

\begin{figure}[!h]
\centering
\includegraphics[width=\linewidth]{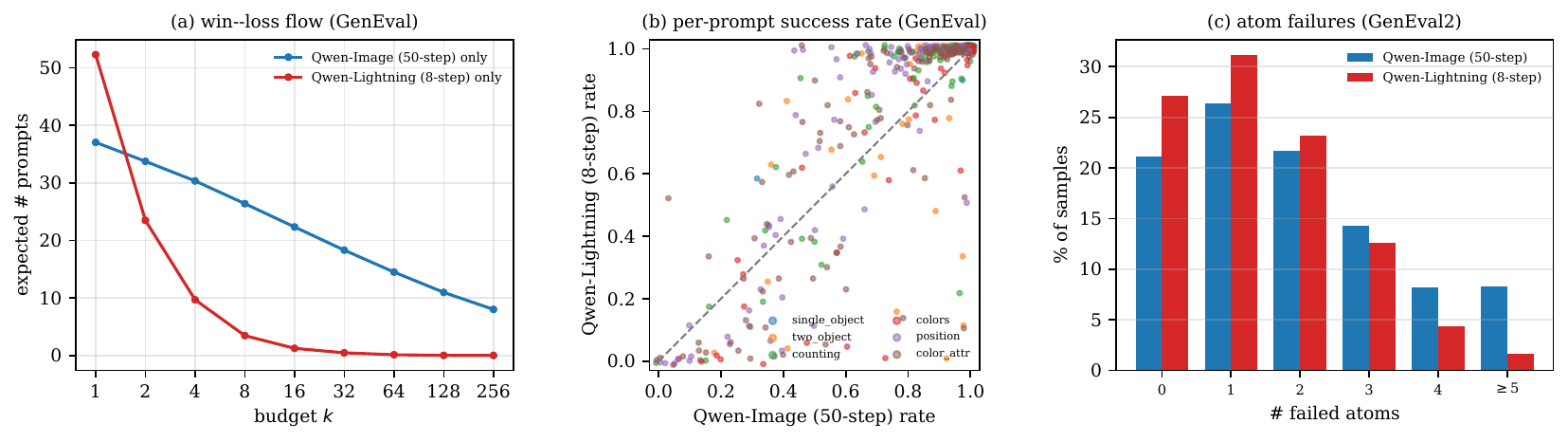}
\caption{\textbf{Per-prompt diagnostics for Qwen-Image and Lightning.}
Panels follow Fig.~\ref{fig:fxd-fam-turbo}.}
\label{fig:fxd-fam-qwen}
\end{figure}

\begin{figure}[!h]
\centering
\includegraphics[width=\linewidth]{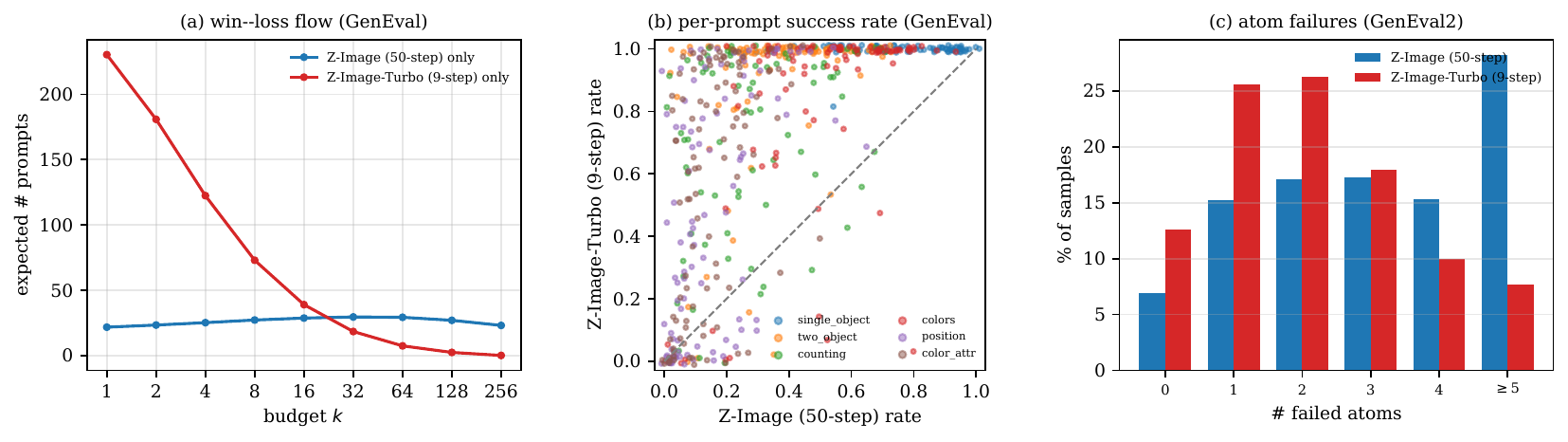}
\caption{\textbf{Per-prompt diagnostics for Z-Image and Z-Image-Turbo.}
Panels follow Fig.~\ref{fig:fxd-fam-turbo}.}
\label{fig:fxd-fam-zimage}
\end{figure}

\begin{figure}[!h]
\centering
\includegraphics[width=\linewidth]{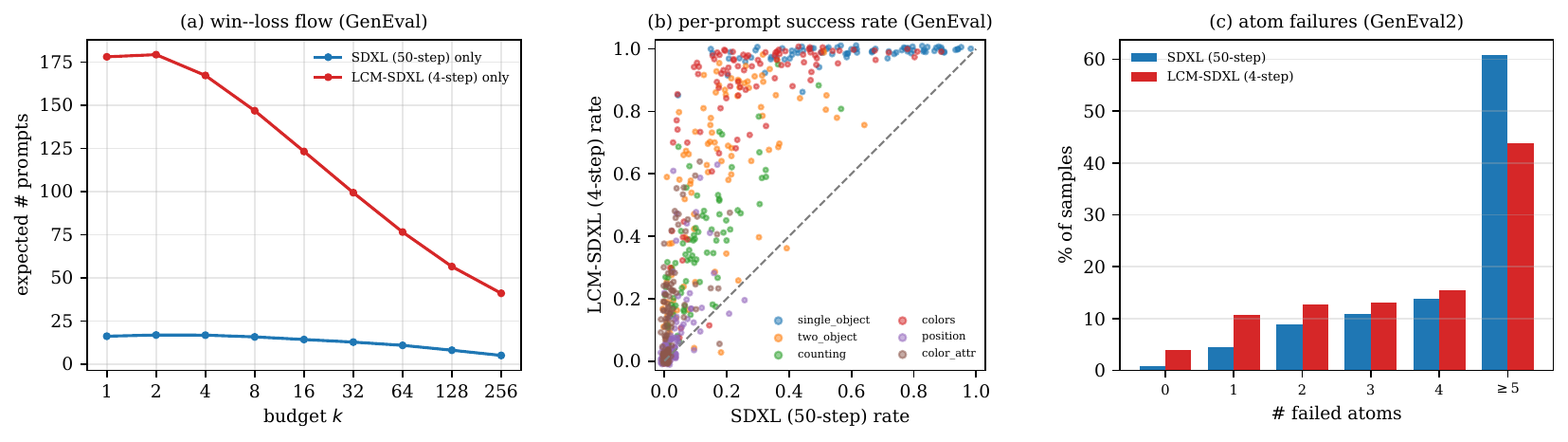}
\caption{\textbf{Per-prompt diagnostics for SDXL and LCM.}
Panels follow Fig.~\ref{fig:fxd-fam-turbo}; LCM retains its advantage across
the measured range.}
\label{fig:fxd-fam-sdxl}
\end{figure}

Figs.~\ref{fig:fxd-fam-turbo}--\ref{fig:fxd-fam-sdxl} show that the aggregate
crossovers reflect a redistribution of success across prompts rather than a
uniform change in correctness. For SD3.5-Turbo, Sana-Sprint,
Qwen-Image-Lightning, and Z-Image-Turbo, distillation produces more student-only
successes at small budgets, but these gains are concentrated on prompts the
student solves repeatedly; as $k$ grows, the teacher continues to recover
prompts outside this set. The GenEval2 failure distributions show that better
sample-level
compositional accuracy can therefore coexist with weaker prompt coverage.
LCM-SDXL behaves differently: its gains persist across prompts and sampling
budgets, consistent with the absence of an aggregate crossover under
consistency distillation.

\FloatBarrier

\section{Additional Controlled FLUX Results}
\label{app:passk_sweeps}

\subsection{Coverage at the Final Checkpoint}

\begin{figure}[!h]
\centering
\includegraphics[width=\linewidth,trim=0 4pt 0 4pt,clip]{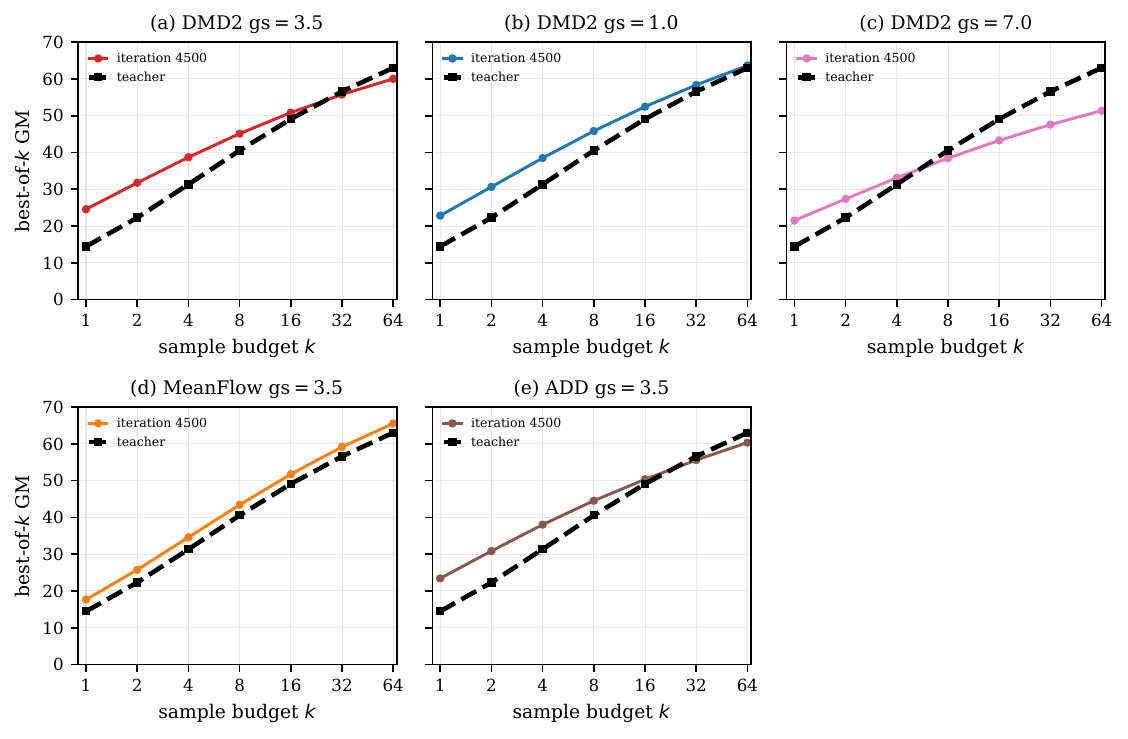}
\caption{\textbf{Best-of-$k$ curves at iteration 4500.}
The top row shows DMD2 with embedded guidance $3.5$, $1$, and $7$; the bottom
row shows MeanFlow and ADD with guidance $3.5$. The dashed black curve is the
shared 50-step FLUX.1-dev teacher.}
\label{fig:passk_sweeps}
\end{figure}

Fig.~\ref{fig:passk_sweeps} compares the best-of-$k$ curves at iteration 4500.
Under embedded guidance $3.5$, DMD2 begins with strong small-budget
performance, but the teacher overtakes as $k$ increases. With embedded
guidance $1$, DMD2 shows a narrowing advantage as $k$ increases, whereas at
guidance $7$ it falls below the teacher from $k{=}8$ onward and trails by more
than $11$ points at $k{=}64$. ADD follows the same pattern as DMD2 at guidance
$3.5$: it leads through $k{=}16$ and falls below the teacher at $k{=}32$ and
$64$. MeanFlow remains above the teacher across the measured range.

\subsection{Per-Prompt Success Rates}

\begin{figure}[!h]
\centering
\includegraphics[width=0.62\linewidth]{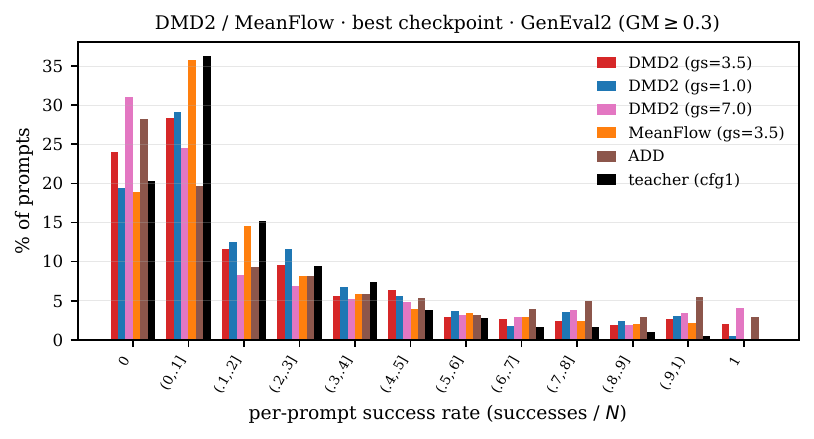}
\caption{\textbf{Per-prompt success rates in the controlled study.}
A sample counts as a success when its GenEval2 geometric-mean score is at
least $0.3$, the cutoff shared with Figs.~\ref{fig:cfg-collapse}
and~\ref{fig:cfg-dist}.}
\label{fig:2x2-dist}
\end{figure}

Fig.~\ref{fig:2x2-dist} shows the per-prompt success-rate distributions at
the selected checkpoints, with success defined by the same cutoff
(Soft-TIFA GM$\geq$0.3) used for all GenEval2 per-prompt diagnostics
(Appendix~\ref{sec:distributions}). Relative to the teacher, the
divergence-based students shift mass toward both extremes: DMD2 (guidance
$3.5$, $7$) and ADD solve more prompts at high rates but also leave more
prompts with zero successes, mirroring the redistribution observed under
strong guidance in Fig.~\ref{fig:cfg-collapse}. MeanFlow instead tracks the
teacher's distribution most closely and leaves the fewest prompts unsolved.
The high-success mass accounts for the strong small-budget scores of DMD2
and ADD, whereas their enlarged zero-success bin limits the gains available
at large $k$, consistent with the teacher overtaking them in
Fig.~\ref{fig:passk_sweeps}.

\FloatBarrier

\section{Evaluator False Positives and the Robustness of \texorpdfstring{Pass@$k$}{Pass@k}}
\label{app:fp}

Unlike deterministic code-execution tests, detector-based evaluators
have nonzero false-positive (FP) rates. If a sample of an unsolved prompt
passes the evaluator with probability $\varepsilon$, the prompt's observed
pass@$k$ contribution saturates toward $1-(1-\varepsilon)^k$ rather than $0$,
so at large budgets FP mass accumulates and compresses differences between
models at the extreme upper tail of the curves. Because every model on a
benchmark is scored by the same evaluator, this inflation is shared:
orderings and crossovers at moderate budgets are far more robust than
absolute values at $k \to n$. This appendix quantifies both effects.

\paragraph{Two FP diagnostics.}
An FP-driven ``solved'' prompt has a signature: its pass@$256$ rests on one
or two passing samples out of $256$. We therefore recompute all comparisons
under two variants. First, an \emph{FP-immune statistic} that requires at
least two successes, which a single false positive cannot satisfy:
\begin{equation}
  \operatorname{pass}_{\geq 2}@k
  =
  \mathbb{E}_{\text{instances}}\!\left[
  1 - \binom{n-c}{k}\Big/\binom{n}{k}
    - c\,\binom{n-c}{k-1}\Big/\binom{n}{k}
  \right].
  \label{eq:passk-ge2}
\end{equation}
Second, for continuous evaluators we sweep the success threshold $\tau$
(success $=$ score $\geq \tau$), since FP scores concentrate near the
threshold.

\begin{table}[htbp]
\centering
\small
\setlength{\tabcolsep}{4.5pt}
\caption{\textbf{Teacher--student crossovers and endpoint gaps survive
FP-immune re-analysis.} For each comparison: the budget at which the
multi-step model first overtakes ($k_\times$) and the $k{=}256$ gap
$\Delta_{256}$ (multi-step minus few-step), under the standard statistic and
the FP-immune $\operatorname{pass}_{\geq 2}@k$ of Eq.~\ref{eq:passk-ge2}.
Success thresholds: VQAScore $\geq 0.8$ (GenAI-Bench), Soft-TIFA GM
$\geq 0.3$ (GenEval2); GenEval is binary. Requiring two successes shifts
each crossover by at most one budget doubling and preserves (GenEval:
enlarges) every gap; SDXL/LCM is the coverage counter-example reported in
the main text and remains one.}
\label{tab:fp-robustness}
\begin{tabular}{llcccc}
\toprule
 & & \multicolumn{2}{c}{standard} & \multicolumn{2}{c}{require $\geq 2$} \\
\cmidrule(lr){3-4}\cmidrule(lr){5-6}
benchmark & comparison & $k_\times$ & $\Delta_{256}$ & $k_\times$ & $\Delta_{256}$ \\
\midrule
GenAI-Bench & SD3.5-L / Turbo & 8 & $+.063$ & 16 & $+.059$ \\
 & Sana / Sprint & 8 & $+.063$ & 16 & $+.063$ \\
 & Qwen-Image / Lightning & 4 & $+.083$ & 8 & $+.072$ \\
 & Z-Image / Turbo & 8 & $+.108$ & 16 & $+.101$ \\
 & SDXL / LCM & 256 & $+.004$ & --- & $-.009$ \\
 & CFG $1$ vs.\ $7$ (SD3.5-L) & 16 & $+.042$ & 32 & $+.042$ \\
\midrule
GenEval & Qwen-Image / Lightning & 2 & $+.014$ & 4 & $+.019$ \\
 & Z-Image / Turbo & 32 & $+.039$ & 64 & $+.044$ \\
\midrule
GenEval2 & Z-Image / Turbo & 8 & $+.156$ & 16 & $+.191$ \\
\bottomrule
\end{tabular}
\end{table}

\paragraph{Robustness of the reported comparisons.}
Table~\ref{tab:fp-robustness} summarizes the re-analysis over all
teacher--student pairs with retained per-sample scores, plus the CFG sweep.
Every crossover reported in the main text survives the FP-immune statistic,
shifting by at most one budget doubling, and every endpoint gap is preserved; on
GenEval the gaps grow, because the few-step students' tails rest on
1--2-hit prompts slightly more than their teachers'. Threshold sensitivity
behaves as the FP model predicts: at $\tau=0.7$ all crossovers occur
earlier, and at the strictest cut ($\tau=0.9$) the upper tail compresses,
delaying the SD3.5-L and Sana crossovers toward the edge of the
measured budget while Qwen-Image ($k_\times{=}8$) and Z-Image
($k_\times{=}64$) persist. The robust finding is therefore the ordering
itself: the few-step student leads at small budgets and its teacher
overtakes at larger ones. The exact crossover budget depends
on how strictly success is defined.

\paragraph{Direct calibration of $\varepsilon$: a null-pair test.}
We measure each evaluator's FP rate directly by scoring images against
\emph{mismatched} prompts: image directory $i$ keeps prompt $i$'s evaluation
program, but contains images generated for prompt $j = (i + \delta) \bmod P$
(a fixed-point-free assignment; $\delta$ coprime with $P$). A pass under this
pairing is counted as a false positive. This measures FPs on unrelated
content; FPs on near-miss images are what the two diagnostics above target.
We score $4$ mismatched pairs per prompt using one model's images:
$800 \times 4$ pairs for GenEval2 Soft-TIFA (Z-Image) and $527 \times 4$ for
GenAI-Bench VQAScore (Qwen-Image). Because the null pairing makes image
content irrelevant to the prompt, $\varepsilon$ is a property of the evaluator
alone and does not depend on which model generated the images.

\begin{table}[htbp]
\centering
\small
\setlength{\tabcolsep}{3.2pt}
\caption{\textbf{Null-pair calibration.} Measured FP rate of each evaluator
and the most it can add to each model's pass@$256$.}
\label{tab:null-pair}
\begin{tabular}{llccc}
\toprule
 & & unsolved & FP & measured \\
evaluator & model & $1-\hat p$ & inflation & $\Delta_{256}$ \\
\midrule
Soft-TIFA GM $\geq 0.3$ & Z-Image & $.011$ & $.001$ [$.005$] & $+.156$ \\
($1/3200$ null passes) & \quad Turbo & $.168$ & $.014$ [$.077$] & \\
\midrule
VQAScore $\geq 0.8$ & SD3.5-L & $.082$ & $0$ [$.036$] & $+.063$ \\
($0/2108$ null passes) & \quad Turbo & $.144$ & $0$ [$.063$] & \\
 & Sana & $.089$ & $0$ [$.039$] & $+.063$ \\
 & \quad Sprint & $.152$ & $0$ [$.067$] & \\
 & Qwen-Image & $.061$ & $0$ [$.027$] & $+.083$ \\
 & \quad Lightning & $.144$ & $0$ [$.063$] & \\
 & Z-Image & $.070$ & $0$ [$.031$] & $+.108$ \\
 & \quad Turbo & $.178$ & $0$ [$.078$] & \\
\bottomrule
\end{tabular}
\end{table}

The point-estimate FP rate is $\hat\varepsilon = 3.1\times10^{-4}$ for
Soft-TIFA ($1/3200$) and $0$ for VQAScore ($0/2108$); the corresponding
pass@$256$ floors for a completely unsolved prompt are
$f = 1-(1-\varepsilon)^{256} = .077$ and $0$, respectively ($.316$ and $.305$
with the one-sided $95\%$ Clopper--Pearson upper bound; for $0/2108$ this is
the rule-of-three bound $1.4\times10^{-3}$). FP inflation in
Table~\ref{tab:null-pair} is the most that FPs can add to a model's
pass@$256$ $\hat p$: $(1-\hat p)f/(1-f)$, where $1-\hat p$ is the fraction
of prompts the model leaves unsolved; bracketed values use the upper-bound
$f$. Rows list the teachers and students of the coverage-loss families of
Table~\ref{tab:fp-robustness}; the gap is given on the teacher's row.

At the measured FP rates, false positives cannot account for the
teacher--student gaps. A false positive can only make an unsolved prompt look
solved, and teachers leave few prompts unsolved (the $1-\hat p$ column). So
even if every false positive went to the teacher and none to the student, the
teacher's pass@$256$ would rise by at most its FP inflation. With
the point estimates this is below the measured gap for every family, and with the
$95\%$ upper bounds in brackets it is too. Moreover, both models share the
same evaluator, and the student has more unsolved prompts, so it gains more
false solves than the teacher.

\end{document}